\documentclass[11pt]{article}
\usepackage[preprint]{acl}
\usepackage{times}
\usepackage{latexsym}
\usepackage[T1]{fontenc}
\usepackage[utf8]{inputenc}
\usepackage{microtype}
\usepackage{inconsolata}
\usepackage{graphicx}
\usepackage{amsmath,amssymb}
\usepackage{booktabs}
\usepackage{xcolor}
\usepackage{url}

\makeatletter
\newcommand{\resdef}[2]{\expandafter\def\csname sl@res@#1\endcsname{#2}}
\newcommand{\res}[1]{\ifcsname sl@res@#1\endcsname\csname sl@res@#1\endcsname\else\textcolor{red}{\textbf{??#1}}\fi}
\makeatother

\resdef{accBankingJeffLargertd}{80.4}
\resdef{accBankingvote}{78.7}
\resdef{accClincJeffLargertd}{92.8}
\resdef{accClincvote}{80.6}
\resdef{accDbpediaJeffLargertd}{92.5}
\resdef{accDbpediavote}{91.1}
\resdef{accOnetJeffLargertd}{68.1}
\resdef{accOnetJevknnFive}{64.1}
\resdef{accOnetJevknntau}{70.1}
\resdef{accOnetRtdMax}{70.1}
\resdef{accOnetRtdMin}{62.0}
\resdef{accOnetbeamThreeMax}{44.4}
\resdef{accOnetbeamThreeMin}{29.7}
\resdef{accOnetvote}{42.6}
\resdef{allFlatVbeamThreeMaxAbs}{31.3}
\resdef{allFlatVbeamThreeMinAbs}{17.1}
\resdef{allFlatVbeamThreeN}{13}
\resdef{allHashVtreeMaxAbs}{33.4}
\resdef{allHashVtreeMinAbs}{2.3}
\resdef{allHashVtreeN}{16}
\resdef{allRtdVbeamThreeMaxAbs}{34.3}
\resdef{allRtdVbeamThreeMinAbs}{16.6}
\resdef{allRtdVbeamThreeN}{16}
\resdef{allRtdVgreedyMaxAbs}{34.8}
\resdef{allRtdVgreedyMinAbs}{19.4}
\resdef{allRtdVvoteN}{24}
\resdef{allRtdVvotePos}{18}
\resdef{allStratVtreeMaxAbs}{33.2}
\resdef{allStratVtreeMinAbs}{6.4}
\resdef{bankDbpedia}{240{,}942}
\resdef{bankRowsDbpediaHash}{23{,}967}
\resdef{bankRowsOnetHash}{4{,}945}
\resdef{callsClincbeamThree}{2}
\resdef{callsDbpediabeamThree}{3}
\resdef{callsOnetbeamThree}{4.4}
\resdef{costBankingflat}{0.068}
\resdef{costBankingknntau}{0.021}
\resdef{costClincflat}{0.102}
\resdef{costClincknntau}{0.019}
\resdef{costDbpediaflat}{0.161}
\resdef{costDbpediaknntau}{0.022}
\resdef{costOnetbeamThree}{0.133}
\resdef{costOnetknntau}{0.034}
\resdef{cov5Onet}{73.7}
\resdef{covTauBanking}{99.0}
\resdef{covTauClinc}{98.2}
\resdef{covTauDbpedia}{99.2}
\resdef{covTauOnet}{86.2}
\resdef{covTauOnetFull}{86.2}
\resdef{covTauOnetHash}{54.2}
\resdef{dBankingHashVFullMaxAbs}{1.5}
\resdef{dBankingHashVFullMinAbs}{4.5}
\resdef{dBankingStratVFullMaxAbs}{0.8}
\resdef{dBankingStratVFullMinAbs}{5.0}
\resdef{dBankingrtdVflatEikosFour}{$+$5.6}
\resdef{dBankingrtdVflatJev}{$+$0.8}
\resdef{dClincHashVFullMaxAbs}{3.7}
\resdef{dClincHashVFullMinAbs}{5.6}
\resdef{dClincStratVFullMaxAbs}{3.0}
\resdef{dClincStratVFullMinAbs}{5.2}
\resdef{dClincrtdVbeamThreeJev}{$+$23.1}
\resdef{dClincrtdVbeamThreeJevHi}{$+$24.9}
\resdef{dClincrtdVbeamThreeJevLo}{$+$21.4}
\resdef{dClincrtdVflatEikosFour}{$+$6.9}
\resdef{dClincrtdVflatJev}{$+$0.7}
\resdef{dClincrtdVvoteMaxAbs}{15.2}
\resdef{dClincrtdVvoteMinAbs}{8.0}
\resdef{dDbpediaHashVFullMinAbs}{0.8}
\resdef{dDbpediaStratVFullMinAbs}{1.0}
\resdef{dDbpediartdVbeamThreeJev}{$+$18.2}
\resdef{dDbpediartdVbeamThreeJevHi}{$+$19.4}
\resdef{dDbpediartdVbeamThreeJevLo}{$+$17.0}
\resdef{dDbpediartdVflatEikosFour}{$+$7.2}
\resdef{dDbpediartdVflatJev}{$+$0.3}
\resdef{dDbpediartdVvoteMaxAbs}{2.2}
\resdef{dOnetHashVFullMaxAbs}{19.6}
\resdef{dOnetHashVFullMinAbs}{27.1}
\resdef{dOnetStratVHashKeptMax}{$-$0.2}
\resdef{dOnetStratVHashKeptMin}{$-$0.8}
\resdef{dOnetStratVHashLostMax}{$+$40.7}
\resdef{dOnetStratVHashLostMin}{$+$34.4}
\resdef{dOnetStratVHashMaxAbs}{4.4}
\resdef{dOnetStratVHashMinAbs}{3.4}
\resdef{dOnetrtdVbeamThreeJev}{$+$25.7}
\resdef{dOnetrtdVbeamThreeJevHi}{$+$27.1}
\resdef{dOnetrtdVbeamThreeJevLo}{$+$24.2}
\resdef{dOnetrtdVbeamThreeMaxAbs}{32.2}
\resdef{dOnetrtdVbeamThreeMinAbs}{25.7}
\resdef{dOnetrtdVflatEikosFour}{$+$4.8}
\resdef{dOnetrtdVflatEikosTwentySeven}{$-$1.1}
\resdef{dOnetrtdVflatStartLux}{$-$2.1}
\resdef{dOnetrtdVvoteMaxAbs}{27.4}
\resdef{dOnetrtdVvoteMinAbs}{19.3}
\resdef{exGold}{bill balance}
\resdef{exGoldDomain}{banking}
\resdef{exKnnFiveRightOf}{all 6}
\resdef{exText}{what is the amount i owe to clay electric}
\resdef{exWrongDomain}{utility}
\resdef{exWrongNOf}{5 of the 6}
\resdef{fanoutOnet}{23}
\resdef{ktauBanking}{11.9}
\resdef{ktauClinc}{10.0}
\resdef{ktauDbpedia}{4.5}
\resdef{ktauOnet}{20.4}
\resdef{labelsBanking}{77}
\resdef{labelsOnet}{1{,}016}
\resdef{latClincbeamThree}{0.79}
\resdef{latClincknntau}{0.40}
\resdef{latDbpediabeamThree}{1.35}
\resdef{latDbpediaknntau}{0.52}
\resdef{latOnetbeamThree}{1.95}
\resdef{latOnetknntau}{0.44}
\resdef{leafClincgreedyMax}{98.7}
\resdef{leafClincgreedyMin}{90.9}
\resdef{leafDbpediagreedyMax}{82.7}
\resdef{leafDbpediagreedyMin}{78.4}
\resdef{leafOnetgreedyMax}{62.7}
\resdef{leafOnetgreedyMin}{50.1}
\resdef{lostLabelsOnetHash}{114}
\resdef{lostQueriesOnetHash}{610}
\resdef{nOpen}{5}
\resdef{noFullN}{6}
\resdef{noFullRtdMax}{92.8}
\resdef{noFullRtdMin}{66.2}
\resdef{noMethodDatasets}{BANKING77, DBpedia, O*NET}
\resdef{noMethodN}{3}
\resdef{noMethodRtdMax}{92.5}
\resdef{noMethodRtdMin}{68.1}
\resdef{probeCap}{255}
\resdef{probeCapN}{10}
\resdef{probeModels}{13}
\resdef{probeOptions}{1{,}016}
\resdef{probeRejectedOf}{all 13}
\resdef{readsClincJeffflat}{11}
\resdef{readsOnetJeffknntau}{2.5}
\resdef{readsOnetflatMax}{42}
\resdef{readsOnetflatMin}{8}
\resdef{rootKidsClinc}{10}
\resdef{rootKidsOnet}{23}
\resdef{routeClincflatMax}{99.0}
\resdef{routeClincflatMin}{92.6}
\resdef{routeClincgreedyMax}{76.4}
\resdef{routeClincgreedyMin}{59.4}
\resdef{routeClincrtdMax}{98.8}
\resdef{routeClincrtdMin}{96.5}
\resdef{routeDbpediagreedyMax}{90.2}
\resdef{routeDbpediagreedyMin}{71.8}
\resdef{routeDbpediartdMax}{98.7}
\resdef{routeDbpediartdMin}{97.9}
\resdef{routeOnetgreedyMax}{69.9}
\resdef{routeOnetgreedyMin}{58.4}
\resdef{routeOnetrtdMax}{84.0}
\resdef{routeOnetrtdMin}{79.6}
\resdef{tokRatioJevMax}{7.4}
\resdef{tokRatioJevMin}{3.2}
\resdef{tokRatioOnetMax}{50.3}
\resdef{tokRatioOnetMin}{36.8}
\resdef{tokRatioWithinMax}{22.9}
\resdef{tokRatioWithinMin}{3.2}
\resdef{withinHashVflatN}{15}
\resdef{withinHashVflatNotNeg}{11}
\resdef{withinHashVfullMinAbs}{5.6}
\resdef{withinRtdVflatMax}{$+$7.2}
\resdef{withinRtdVflatMin}{$-$0.5}
\resdef{withinRtdVflatN}{15}
\resdef{withinRtdVflatPos}{9}

\newcommand{\taumass}{0.95}
\newcommand{\minanswered}{95\%}  %
\newcommand{\bmk}{1.5}
\newcommand{\bmb}{0.75}

\newcommand{\Choice}{\texttt{choice}}

\hypersetup{
  pdftitle={AdaMenu: Query-Adaptive Menus for Decision Models over Large Label Spaces},
  pdfauthor={Jiangze Yan, Yi Shen, Jieyun Huang, Wenjing Zhang, Zhaoxiang Liu, Ning Wang, Kai Wang, Shiguo Lian}
}

\title{AdaMenu: Query-Adaptive Menus for Decision Models\\over Large Label Spaces}

\author{
  \bfseries Jiangze Yan\textsuperscript{1,2} \quad Yi Shen\textsuperscript{1,2} \quad Jieyun Huang\textsuperscript{1,2} \quad Wenjing Zhang\textsuperscript{1,2} \\
  \bfseries Zhaoxiang Liu\textsuperscript{1,2} \quad Ning Wang\textsuperscript{1,2} \quad Kai Wang\textsuperscript{1,2} \quad Shiguo Lian\textsuperscript{1,2} \\[4pt]
  \textsuperscript{1}Unicom Data Intelligence, China Unicom \\
  \textsuperscript{2}Data Science \& Artificial Intelligence Research Institute, China Unicom \\[2pt]
  \texttt{\{yanjz17, sheny73, liansg\}@chinaunicom.cn}
}

\begin{document}
\maketitle
\begin{abstract}
Decision models such as Jev return probabilities over caller-defined options and are increasingly used for
classification. Large label spaces challenge both accuracy and scalability: larger menus can impair decisions,
increase input costs, and exceed option limits. Hierarchical classification reduces menu size but introduces
serial calls and error propagation. We propose \emph{AdaMenu}, which builds a compact candidate menu from
retrieved labeled examples and makes a single decision-model request. Its candidate labels adapt to the query,
and its default setting also adjusts menu size using aggregated retrieval scores. We evaluate AdaMenu with
open and closed decision models on four label sets containing up to \res{labelsOnet} labels. It outperforms
TypeSafe's beam-search baseline on all \res{allRtdVbeamThreeN} model and label-set pairs where both methods run,
improving accuracy by \res{allRtdVbeamThreeMinAbs}--\res{allRtdVbeamThreeMaxAbs} percentage points.
Relative to the full menu on the three smaller label sets, accuracy differences range from
\res{withinRtdVflatMin} to \res{withinRtdVflatMax} percentage points, while input token usage falls by factors of
\res{tokRatioWithinMin}--\res{tokRatioWithinMax}. AdaMenu also runs when option limits exclude the full menu,
including settings where the hierarchical baseline is unavailable.
\end{abstract}

\section{Introduction}
\label{sec:intro}

\begin{figure}[t]
\centering
\includegraphics[width=\columnwidth]{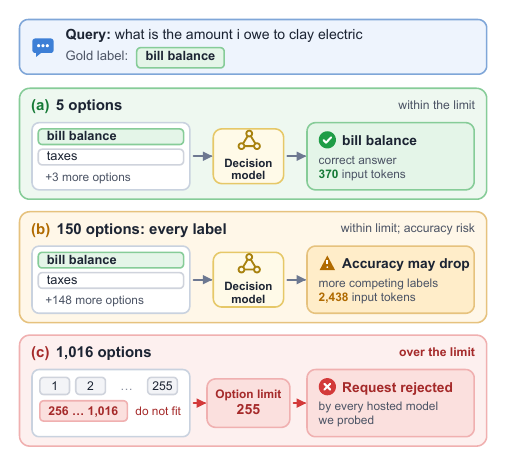}
\caption{Large menus challenge accuracy and scalability.
(a)~A small menu keeps input cost low.
(b)~More options can reduce accuracy and increase input cost, even within the option limit.
(c)~Menus exceeding the limit are rejected.}
\label{fig:teaser}
\end{figure}

A \emph{decision model} answers a typed question. Given a piece of text and a \Choice{} question
whose options the caller defines, it returns a probability for every option, with no generated text to parse. This class
of models was introduced with Jev \citep{almeida2026introducing}, and open models have followed.
Decision models are fast and cheap, and much of
their use is plain classification, with the labels as the options: assigning intent, product or topic labels to
input text.

Large label spaces challenge both accuracy and scalability (Figure~\ref{fig:teaser}). First, adding
candidate options can reduce classification accuracy even when the correct label remains available
\citep{rafe2026benchmarking}; changes to the option set can also alter the model's relative judgments among
candidate labels \citep{rajput2026hosted}. Second, every call reads the menu's options, so input cost increases
with menu size. In addition, a model's option limit constrains the size of the full label set it can accept.
Hierarchical classification along a label tree reduces the number of options at each decision, but requires
multiple calls and can propagate early routing errors.

We propose \emph{AdaMenu} (Figure~\ref{fig:method}), which retrieves labeled examples to construct a compact
candidate menu and classifies each query with a single decision-model request. The candidate labels vary with
the query, and the default setting also adapts their number. This focuses the model on plausible labels,
reduces the cost of reading the full label set and avoids traversing a predefined label tree.

We evaluate AdaMenu on open and closed decision models across four label sets.

Our contributions are as follows.

\begin{itemize}
\item \textbf{Problem analysis.} We examine how menu construction affects accuracy and input costs in large
  label spaces, characterize option limits, and analyze routing errors in the hierarchical
  classification approach (\S\ref{sec:within}, \S\ref{sec:why}).
\item \textbf{Method.} We propose AdaMenu, which constructs a compact, query-adaptive candidate menu from retrieved
  examples and classifies each query with a single decision-model request, without additional model training
  (\S\ref{sec:method}).
\item \textbf{Classification beyond the option limit.} AdaMenu handles classification tasks whose full label sets
  exceed the model's option limit with a single request, including settings where the hierarchical baseline is
  also unavailable (\S\ref{sec:beyond}).
\item \textbf{Accuracy and cost within the limit.} Where the full menu is accepted on the three smaller label sets,
  AdaMenu achieves competitive accuracy with fewer input tokens (\S\ref{sec:within}). It also outperforms
  TypeSafe's beam-search baseline on all \res{allRtdVbeamThreeN} comparable model and label-set pairs,
  using one request instead of several serial requests (\S\ref{sec:vs-hier}).
\end{itemize}

\section{Related Work}
\label{sec:related}

Open instruction-tuned models can be adapted to return option probabilities \citep{geuter2026releasing}, while
architectural work addresses sensitivity to option order and interactions among options
\citep{teng2026prism,li2026jeva,antunes2026narrow}. Adding options changes Jev's relative probabilities
\citep{rajput2026hosted}, and option labels can outweigh written definitions in open-weight models
\citep{azizi2026labels}. On CLINC150, \citet{rafe2026benchmarking} report lower accuracy as Jev's menu grows
while retaining the gold label. \citet{li2026jevcascade} report higher accuracy with curated candidate menus
than with the full label set.

Top-down hierarchical classification reduces menu size through local decisions along a label tree
\citep{silla2011survey}. TypeSafe, Jev's developer, recommends using the decision model for these choices
with beam search \citep{typesafe2026hierarchical}. \citet{rafe2026benchmarking} trace errors in coarse-to-fine
CLINC150 classification to the initial domain decision; \citet{li2026jevcascade} report accuracy losses from
coarse-to-fine classification on TREC. We use TypeSafe's procedure as the hierarchical baseline and analyze
its routing errors.

Candidate retrieval followed by more expensive scoring is used in entity linking \citep{wu2020scalable} and
extreme multi-label classification with language models \citep{doosterlinck2024incontext}. Retrieval also supports
different forms of prediction: nearest-neighbor language models augment token predictions with datastore
neighbors \citep{khandelwal2020generalization}, while $k$NN Prompting classifies using neighbors in representations
derived from language-model outputs \citep{xu2023knn}. For in-context learning with many labels, retrieval selects
demonstrations that expose a subset of the label space \citep{milios2023incontext}. AdaMenu uses retrieved examples
to construct a candidate menu for a decision-model API request.

\section{Method}
\label{sec:method}

\begin{figure*}[t]
\centering
\includegraphics[width=\textwidth]{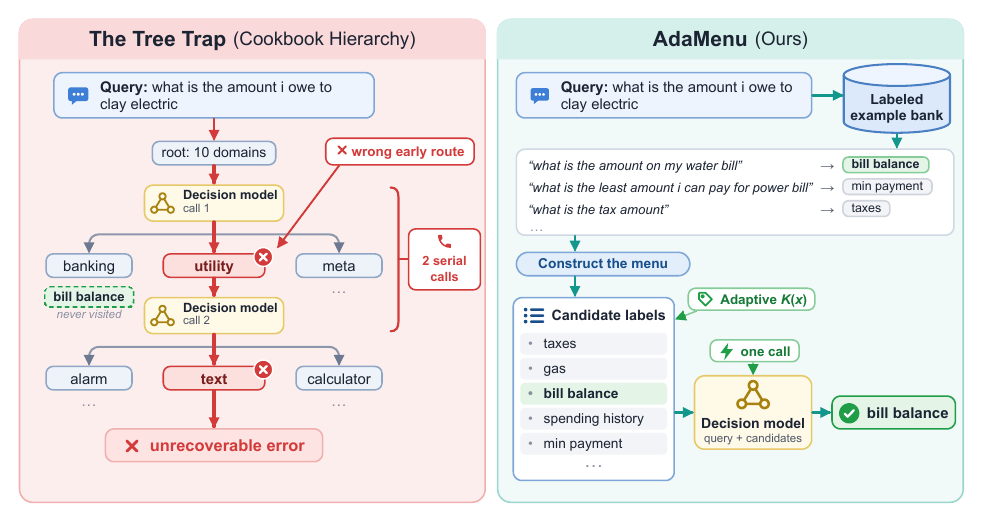}
\caption{Greedy label-tree traversal versus AdaMenu, on the query of Figure~\ref{fig:teaser}.
\emph{Left:} successive decisions along the tree can exclude the correct branch early.
\emph{Right:} retrieval constructs an adaptive candidate menu (partially shown), followed by a single
decision-model request.}
\label{fig:method}
\end{figure*}

A decision model receives a query $x$ and a caller-defined menu $O$, and returns a probability
$p(o \mid x,O)$ for each option $o \in O$. Given a label set $\mathcal{Y}$, the direct classifier uses the
full menu $O=\mathcal{Y}$, when the model accepts it, and predicts
$\hat y_{\mathrm{full}}(x)=\arg\max_{y\in\mathcal{Y}}p(y\mid x,\mathcal{Y})$.

AdaMenu assumes a bank $\mathcal{B}=\{(x_i,y_i)\}$ of labeled examples, where $x_i$ is the text of the $i$th
example and $y_i\in\mathcal{Y}$ is its label. The bank can come from a training set or a log of past decisions.
AdaMenu constructs a candidate menu $C(x)\subseteq\mathcal{Y}$ for each query and uses one decision-model
request to classify it.
The workflow consists of three steps (Figure~\ref{fig:method}, right):
\begin{enumerate}
\item \textbf{Retrieve.} Score the bank examples against $x$ and let $N_M(x)$ contain the indices of up to
  $M$ highest-scoring examples, with scores $s(x,x_i)$. Here $M$ counts examples; the menu size $K(x)$ counts
  distinct candidate labels. The retriever and its settings are specified in \S\ref{sec:setup}.
\item \textbf{Construct the menu.} For each label $y$, define its query-dependent score $r_x(y)$ by aggregating
  positive retrieval scores:
  \begin{equation}
  r_x(y)=\sum_{\substack{i\in N_M(x),\ y_i=y\\s(x,x_i)>0}}s(x,x_i).
  \label{eq:label-score}
  \end{equation}
  Rank labels with positive $r_x(y)$ in descending order. If necessary, extend this list using later
  retrieved examples and the remaining labels, as specified in Appendix~\ref{app:candidates}. Let
  $\pi_x=(y_{(1)},\ldots,y_{(m_x)})$ be this ordered list before truncation, where $m_x$ is its length and
  $y_{(j)}$ is its $j$th label. The list contains every positive-score label. Let
  $R_x=\sum_{y\in\mathcal{Y}}r_x(y)$ be the total label score and $0<\tau\leq1$ the target fraction of that total.
  When $R_x>0$, the shortest prefix meeting this target has length
  \begin{equation}
  \widetilde K(x)=\min\left\{k:\sum_{j=1}^{k}r_x(y_{(j)})\geq\tau R_x\right\},
  \label{eq:prefix-length}
  \end{equation}
  where $k$ ranges over the integer prefix lengths $1,\ldots,m_x$. Let $K_{\min}$ and $K_{\max}$ be the
  configured lower and upper menu-size bounds, with $1\leq K_{\min}\leq K_{\max}$.
  If $R_x=0$, set $\widetilde K(x)=K_{\max}$. Clamp $\widetilde K(x)$ to $[K_{\min},K_{\max}]$, then cap it
  at the available count $m_x$ to obtain $K(x)$. The candidate menu consists of the first $K(x)$ labels,
  $C(x)=\{y_{(1)},\ldots,y_{(K(x))}\}$.
  The upper bound can truncate the selected prefix, so the retained fraction of total score may be below
  $\tau$. The threshold controls cumulative retrieval score; candidate coverage is measured empirically.
  Fixed-size variants use a chosen positive integer budget $K\leq K_{\max}$ and take the first $\min(K,m_x)$ labels
  from the same query-dependent order. The chosen menu must fit the model's
  request-level option limit; model-specific settings are given in \S\ref{sec:setup}.
\item \textbf{Decide.} Make one \Choice{} request over $C(x)$ with shuffled options and the instruction
  \emph{``Which category best matches this document?''} The model receives only the query and candidate labels,
  without retrieval scores or retrieved examples. Its predicted label is
  \begin{equation}
  \hat y(x)=\arg\max_{y\in C(x)}p(y\mid x,C(x)).
  \label{eq:adamenu-prediction}
  \end{equation}
\end{enumerate}

Because $\hat y(x)\in C(x)$, accuracy is bounded by coverage of the gold label $y^\star$:
\begin{equation}
\Pr\!\left(\hat y(x)=y^\star\right)\leq\Pr\!\left(y^\star\in C(x)\right).
\label{eq:coverage-bound}
\end{equation}
High coverage alone does not ensure high accuracy: the composition of the menu also affects the model's
ability to distinguish its candidates. Candidate construction must balance retaining the correct label
against distracting options and input cost. The retriever need not rank the correct label first.
We report coverage alongside accuracy (\S\ref{sec:parts}).

\section{Experiments}
\label{sec:experiments}

\subsection{Setup}
\label{sec:setup}

We test one closed and \res{nOpen} open decision models (Table~\ref{tab:models}). Jev-1.14
\citep{almeida2026introducing} is queried through OpenRouter's decision endpoint. The open models, Jeff-1.1-Qwen-4B
and Jeff-1.0-Large \citep{geuter2026releasing}, StartLux-4B \citep{startlux2026decision}, and Eikos-4B and Eikos-27B
\citep{vicentino2026eikos}, run on our own GPUs, with each vendor's prompt, readout and handling of long menus
re-implemented from its public serving code (Appendix~\ref{app:models}). A method is available for a model
only if the model answers at least \minanswered{} of the queries, which excludes menus longer than it accepts;
unanswered queries count as errors.

We count requests at the decision-model interface separately from internal forward passes (reads). The fixed-size
$K{=}10$ menu fits in one request for every model tested and within the single-pass capacity of every open
model. An adaptive menu can require multiple internal reads on some open models;
Appendix~\ref{app:models} details the serving behavior and the model-specific settings.

\begin{table}[t]
\centering\footnotesize\setlength{\tabcolsep}{3.5pt}

\begin{tabular}{@{}lrrl@{}}
\toprule
Model & One read & Most & Longer menus \\
\midrule
Jev-1.14 (closed) & -- & 255 & rejected \\
Jeff-1.1-Qwen-4B & 16 & 256 & groups + final read \\
StartLux-4B & 25 & any & knock-out \\
Eikos-4B & 100 & any & tournament \\
Eikos-27B & 160 & any & tournament \\
Jeff-1.0-Large & 16 & 16 & rejected \\
\bottomrule
\end{tabular}

\caption{The decision models and how they read a menu: options in one forward pass (One read), most options in one
question (Most), and what happens to a longer menu. Jev-1.14 is hosted; the open models are served by us.}
\label{tab:models}
\end{table}

Four English label sets span \res{labelsBanking} to \res{labelsOnet} labels (Table~\ref{tab:data}). In each, the bank
that AdaMenu retrieves from is disjoint from the evaluation queries. \textbf{BANKING77} \citep{casanueva2020efficient}
has no label tree; we evaluate on a hash-selected tenth of its training set and use the rest as the bank.
\textbf{CLINC150} \citep{larson2019evaluation}: all in-scope validation queries; the bank is the in-scope training
set; the tree is the dataset's grouping of the intents into \res{rootKidsClinc} domains. \textbf{DBpedia}
\citep{lehmann2015dbpedia} in the three-level version of \citet{ofer2019dbpedia}, from its Hugging Face copy
\citep{dbpediaclasses}: test articles selected by hash; the bank is the training split. \textbf{O*NET}: job titles
reported by workers and experts in the O*NET 31.0 database \citep{onet2026database}, each listed under exactly one
occupation; the bank is the database's table of alternate job titles, without any title that equals an evaluation
query; the tree is the occupational classification \citep{bls2018soc} from \res{rootKidsOnet} major groups down to the
occupations, with single-child levels collapsed.

\begin{table}[t]
\centering\footnotesize\setlength{\tabcolsep}{3pt}

\begin{tabular}{@{}lrrcc@{}}
\toprule
Data & Queries & Labels & Depth & Fan-out \\
\midrule
BANKING77 & 1{,}033 & 77 & -- & -- \\
CLINC150 & 3{,}000 & 150 & 2 & 15 \\
DBpedia & 5{,}000 & 219 & 3 & 30 \\
O*NET & 5{,}329 & 1{,}016 & 5 & 23 \\
\bottomrule
\end{tabular}

\caption{Label sets. Depth: levels of the label tree. Fan-out: the
most children of any node, i.e., the largest menu used by the hierarchical baseline.}
\label{tab:data}
\end{table}

For retrieval, we use BM25 \citep{robertson2009probabilistic} over lowercased words and adjacent word pairs,
with $k_1=\bmk$ and $b=\bmb$, retrieving $M=50$ examples. The adaptive menu uses $\tau=\taumass$,
$K_{\min}=2$ and $K_{\max}=26$. Deterministic ranking and fallback rules are detailed in
Appendix~\ref{app:candidates}.

We compare the following methods using the same query text. Every menu is shuffled with a fixed seed per query
and method, shared across models.
\begin{itemize}
\item \textbf{AdaMenu} (ours, \S\ref{sec:method}), with adaptive menu sizes; $K{=}10$ for
  Jeff-1.0-Large. Fixed $K \in \{5, 10, 20\}$ are variants (\S\ref{sec:parts}).
\item \textbf{Full menu}: one \Choice{} question over all labels in a single request, with the same instruction as AdaMenu.
\item \textbf{Hierarchical baseline (Tree)}: TypeSafe's cookbook procedure \citep{typesafe2026hierarchical}.
  At each node, it asks a \Choice{} question over the children, with the instruction \emph{``Which direct child category best matches this
  document?''} and the child's name as its description. \emph{Greedy} follows the most probable child; \emph{beam-3}
  keeps the three paths with the highest geometric mean of edge probabilities, with one round of requests per level.
\end{itemize}
To isolate the decision step, we also report AdaMenu's \emph{top-1 kNN candidate}, its top-ranked label without a model
request (\S\ref{sec:parts}).

Accuracy is exact match with the gold label. Cost is the number of input tokens per query, summed over all requests
or reads of a method, the price paid for Jev-1.14, and the number of serial requests; Jev-1.14's wall-clock latency
includes retrieval. For each model and label set (a \emph{pair}), we compare two methods by the paired difference in
accuracy with a 95\% percentile bootstrap over queries (10{,}000 resamples, fixed seed); a difference is significant when its
interval excludes zero.

\begin{table*}[t]
\centering\small\setlength{\tabcolsep}{1.8pt}

\begin{tabular}{@{}l rrr rrrr rrrr rrrr@{}}
\toprule
 & \multicolumn{3}{c}{BANKING77} & \multicolumn{4}{c}{CLINC150} & \multicolumn{4}{c}{DBpedia} & \multicolumn{4}{c}{O*NET} \\
\cmidrule(lr){2-4}\cmidrule(lr){5-8}\cmidrule(lr){9-12}\cmidrule(l){13-16}
Model & Full & Split & AdaMenu & Full & Split & Tree & AdaMenu & Full & Split & Tree & AdaMenu & Full & Split & Tree & AdaMenu \\
\midrule
Jev-1.14 & 79.3 & -- & \textbf{80.1} & 92.3 & -- & 69.9 & \textbf{93.0} & 93.0 & -- & 75.1 & \textbf{93.3} & -- & -- & 44.4 & \textbf{70.1} \\
Jeff-1.1-Qwen-4B & -- & 78.3 & \textbf{80.8} & -- & 90.8 & 68.8 & \textbf{92.2} & -- & 91.8 & 72.0 & \textbf{93.1} & -- & -- & 37.9 & \textbf{66.2} \\
StartLux-4B & -- & 89.2 & \textbf{89.6} & -- & \textbf{96.3} & 79.2 & 95.8 & -- & 91.5 & 73.1 & \textbf{92.8} & -- & \textbf{69.5} & 38.2 & 67.4 \\
Eikos-4B & 70.8 & -- & \textbf{76.4} & -- & 81.7 & 60.0 & \textbf{88.6} & -- & 83.9 & 56.8 & \textbf{91.0} & -- & 57.2 & 29.7 & \textbf{62.0} \\
Eikos-27B & 77.1 & -- & \textbf{79.6} & \textbf{92.7} & -- & 65.8 & 92.2 & -- & 91.5 & 71.2 & \textbf{93.2} & -- & \textbf{68.0} & 40.1 & 66.9 \\
Jeff-1.0-Large & -- & -- & \textbf{80.4} & -- & -- & 72.0 & \textbf{92.8} & -- & -- & -- & \textbf{92.5} & -- & -- & -- & \textbf{68.1} \\
\midrule
Top-1 kNN candidate &  &  & 78.7 &  &  &  & 80.6 &  &  &  & 91.1 &  &  &  & 42.6 \\
\bottomrule
\end{tabular}

\caption{Accuracy (\%) of the full menu, the hierarchical baseline (Tree, beam 3) and AdaMenu; best per model and label set
in bold. Full: the model reads all labels in one pass. Split: its serving code splits them into groups read in
several passes, within one request. --: not available for the model. AdaMenu uses adaptive menu sizes, except $K{=}10$
for Jeff-1.0-Large. Top-1 kNN candidate: AdaMenu without the model request (an ablation). Every method and model:
Table~\ref{tab:allarms}.}
\label{tab:main}
\end{table*}

\begin{figure*}[t]
\centering
\includegraphics[width=\textwidth]{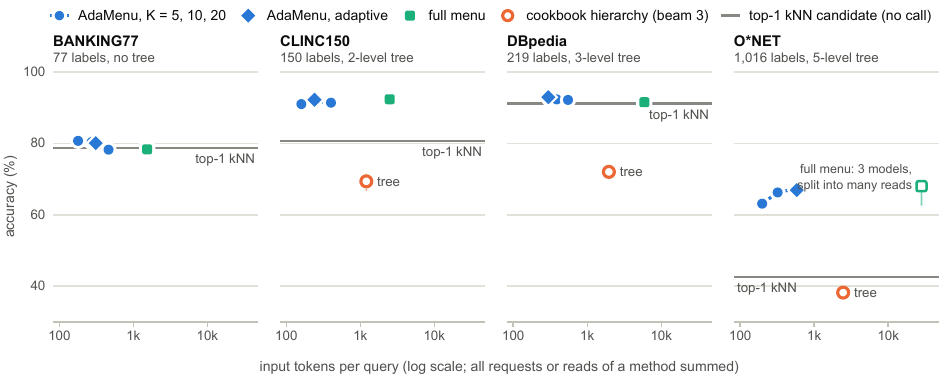}
\caption{Accuracy versus input tokens per query (log-scale x-axis). Points show medians over the models available
for each method; vertical bars show the interquartile range of accuracy.}
\label{fig:tradeoff}
\end{figure*}

\subsection{Within the limit: AdaMenu versus the full menu}
\label{sec:within}

Over the \res{withinRtdVflatN} pairs of BANKING77, CLINC150 and DBpedia where the model accepts the full menu (Full or
Split in Table~\ref{tab:main}), AdaMenu is
significantly more accurate on \res{withinRtdVflatPos} and not significantly different on the rest; the differences
range from \res{withinRtdVflatMin} to \res{withinRtdVflatMax} points (Table~\ref{tab:main}). For Jev-1.14 they are
\res{dBankingrtdVflatJev}, \res{dClincrtdVflatJev} and \res{dDbpediartdVflatJev} points. Eikos-4B shows the largest
gains (\res{dBankingrtdVflatEikosFour}, \res{dClincrtdVflatEikosFour} and \res{dDbpediartdVflatEikosFour} points).

Jeff-1.1-Qwen-4B, for example, reads the full CLINC150 menu in \res{readsClincJeffflat} passes
(Split in Table~\ref{tab:main}).
AdaMenu reduces input token usage relative to the full menu by factors of \res{tokRatioJevMin}--\res{tokRatioJevMax}
for Jev-1.14 and up to \res{tokRatioWithinMax} for the open models
(Figure~\ref{fig:tradeoff}). For Jev-1.14 the price per 1{,}000 queries drops from USD \res{costBankingflat},
\res{costClincflat} and \res{costDbpediaflat} to \res{costBankingknntau}, \res{costClincknntau} and
\res{costDbpediaknntau} (Table~\ref{tab:cost}). Jeff-1.0-Large, which accepts none of the full menus, reaches
\res{accBankingJeffLargertd}\%, \res{accClincJeffLargertd}\% and \res{accDbpediaJeffLargertd}\% with AdaMenu.

\begin{table}[t]
\centering\footnotesize\setlength{\tabcolsep}{2.5pt}

\begin{tabular}{@{}l rrr rrr@{}}
\toprule
 & \multicolumn{2}{c}{Full} & \multicolumn{2}{c}{Tree} & \multicolumn{2}{c}{AdaMenu} \\
\cmidrule(lr){2-3}\cmidrule(lr){4-5}\cmidrule(l){6-7}
Data & Tokens & USD & Tokens & USD & Tokens & USD \\
\midrule
BANKING77 & 1{,}621 & 0.068 & -- & -- & 508 & 0.021 \\
CLINC150 & 2{,}438 & 0.102 & 1{,}432 & 0.060 & 441 & 0.019 \\
DBpedia & 3{,}823 & 0.161 & 2{,}255 & 0.095 & 514 & 0.022 \\
O*NET & -- & -- & 3{,}160 & 0.133 & 804 & 0.034 \\
\bottomrule
\end{tabular}

\caption{Cost of Jev-1.14 per method: Tokens are input tokens per query (all requests); USD is the price per 1{,}000 queries.
Tree: the hierarchical baseline with beam 3. --: no supplied label tree (BANKING77) or menu rejected (O*NET).}
\label{tab:cost}
\end{table}

\subsection{AdaMenu versus the hierarchical baseline}
\label{sec:vs-hier}

\begin{figure}[t]
\centering
\includegraphics[width=\columnwidth]{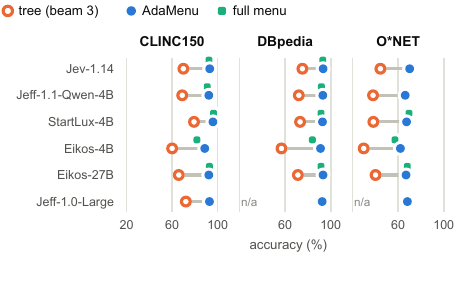}
\caption{Per-model accuracy of the hierarchical baseline (beam 3), AdaMenu and, where the model accepts it, the full menu. n/a: the
tree has nodes with more children than Jeff-1.0-Large reads.}
\label{fig:permodel}
\end{figure}

AdaMenu outperforms the hierarchical baseline with beam 3 on all \res{allRtdVbeamThreeN} pairs where both run,
by \res{allRtdVbeamThreeMinAbs} to
\res{allRtdVbeamThreeMaxAbs} points, and greedy descent by \res{allRtdVgreedyMinAbs} to \res{allRtdVgreedyMaxAbs};
every difference is significant. For Jev-1.14 the gain is \res{dClincrtdVbeamThreeJev} points on CLINC150 (95\% CI
\res{dClincrtdVbeamThreeJevLo} to \res{dClincrtdVbeamThreeJevHi}) and \res{dDbpediartdVbeamThreeJev} on DBpedia
(\res{dDbpediartdVbeamThreeJevLo} to \res{dDbpediartdVbeamThreeJevHi}). The hierarchical baseline also trails the full menu
wherever a model runs it, by \res{allFlatVbeamThreeMinAbs} to \res{allFlatVbeamThreeMaxAbs}
points on all \res{allFlatVbeamThreeN} comparable pairs (Table~\ref{tab:main}, Figure~\ref{fig:permodel}).
For Jev-1.14, the hierarchical baseline uses \res{callsClincbeamThree} and \res{callsDbpediabeamThree} serial requests and takes
\res{latClincbeamThree} and \res{latDbpediabeamThree}\,s per query, against one request and
\res{latClincknntau} and \res{latDbpediaknntau}\,s for AdaMenu.

\subsection{Routing errors in the hierarchical baseline}
\label{sec:why}

We split each prediction into a route and a leaf: does it lie under the gold label's top-level node, and if so, is it
the gold label (Table~\ref{tab:routing})? On CLINC150, the greedy hierarchical baseline first routes
\res{routeClincgreedyMin}--\res{routeClincgreedyMax}\% of queries to the gold domain, while the full menu's predictions
fall in it for \res{routeClincflatMin}--\res{routeClincflatMax}\% and AdaMenu's for
\res{routeClincrtdMin}--\res{routeClincrtdMax}\%. Once in the right domain, greedy descent finds the gold intent for
\res{leafClincgreedyMin}--\res{leafClincgreedyMax}\% of queries. Most of the greedy baseline's errors on CLINC150
occur at the first routing step. CLINC150 files \emph{\res{exGold}} under \emph{\res{exGoldDomain}} and puts timers,
calculators and text messages under \emph{utility}; asked about ``\res{exText}'', \res{exWrongNOf} models step into
\emph{\res{exWrongDomain}} and never see the gold intent, while AdaMenu's adaptive menu retains it
(Figure~\ref{fig:method}). With the fixed-size menu of Figure~\ref{fig:teaser}a,
\res{exKnnFiveRightOf} models choose the gold intent.

Errors also accumulate below the first level on DBpedia and O*NET. On DBpedia the first step routes
\res{routeDbpediagreedyMin}--\res{routeDbpediagreedyMax}\% of queries correctly (AdaMenu: \res{routeDbpediartdMin}--\res{routeDbpediartdMax}\%), and the errors compound over
the levels below: among correctly routed queries, greedy descent reaches the gold class for only
\res{leafDbpediagreedyMin}--\res{leafDbpediagreedyMax}\%. On O*NET, with five levels, the first step is right for
\res{routeOnetgreedyMin}--\res{routeOnetgreedyMax}\% (AdaMenu: \res{routeOnetrtdMin}--\res{routeOnetrtdMax}\%) and the
rest of the walk for \res{leafOnetgreedyMin}--\res{leafOnetgreedyMax}\%.

\subsection{Classification beyond the option limit}
\label{sec:beyond}

A model cannot take the full menu on \res{noFullN} model and label-set pairs: Jev-1.14 and
Jeff-1.1-Qwen-4B on O*NET, and Jeff-1.0-Large on every label set (Table~\ref{tab:models}). On
\res{noMethodN} of them (Jeff-1.0-Large on \res{noMethodDatasets}), the hierarchical baseline is also unavailable:
BANKING77 supplies no label tree, while DBpedia and O*NET have nodes that exceed the model's option limit.
AdaMenu runs on all \res{noFullN} pairs with one request and reaches
\res{noFullRtdMin}--\res{noFullRtdMax}\% accuracy, and \res{noMethodRtdMin}--\res{noMethodRtdMax}\% on the
\res{noMethodN} where neither baseline is available (Table~\ref{tab:main}).

On O*NET, none of the hosted models we probed accepts the full menu (Figure~\ref{fig:teaser}c). The hierarchical
baseline reaches \res{accOnetbeamThreeMin}--\res{accOnetbeamThreeMax}\% accuracy on the five-level label tree
(Table~\ref{tab:main}). AdaMenu answers every query with one request and reaches \res{accOnetRtdMin}--\res{accOnetRtdMax}\%,
between \res{dOnetrtdVbeamThreeMinAbs} and \res{dOnetrtdVbeamThreeMaxAbs} points above the hierarchical baseline
on every model where both methods run (all
significant; Jev-1.14: \res{dOnetrtdVbeamThreeJev}, 95\% CI \res{dOnetrtdVbeamThreeJevLo} to
\res{dOnetrtdVbeamThreeJevHi}). Jeff-1.0-Large cannot run the hierarchical baseline, since the root has
\res{fanoutOnet} children, yet with AdaMenu it reaches \res{accOnetJeffLargertd}\%, second only to Jev-1.14.

For Jev-1.14, the hierarchical baseline costs USD \res{costOnetbeamThree} per 1{,}000 queries and averages
\res{callsOnetbeamThree} serial requests and \res{latOnetbeamThree}\,s per query (Table~\ref{tab:cost}). AdaMenu costs
USD \res{costOnetknntau} per 1{,}000 queries, uses one request per query and averages \res{latOnetknntau}\,s.

Three open models accept the full O*NET menu through a knock-out over groups of labels, with
\res{readsOnetflatMin}--\res{readsOnetflatMax} reads per query on average (Split in Table~\ref{tab:main}).
Relative to these split full menus, AdaMenu changes accuracy by \res{dOnetrtdVflatStartLux} points for StartLux-4B,
\res{dOnetrtdVflatEikosTwentySeven} for Eikos-27B and \res{dOnetrtdVflatEikosFour} for Eikos-4B, while reducing
input token usage by factors of \res{tokRatioOnetMin}--\res{tokRatioOnetMax}.

\subsection{What each part of AdaMenu contributes}
\label{sec:parts}

The top-1 kNN candidate alone is right for \res{accBankingvote}\%, \res{accClincvote}\%, \res{accDbpediavote}\% and
\res{accOnetvote}\% of queries (Table~\ref{tab:main}). Letting the model decide among the candidates is significantly better
on \res{allRtdVvotePos} of \res{allRtdVvoteN} pairs and not significantly different on the rest. It matters most where retrieval alone
is weak: \res{dClincrtdVvoteMinAbs} to \res{dClincrtdVvoteMaxAbs} points on CLINC150 and \res{dOnetrtdVvoteMinAbs} to
\res{dOnetrtdVvoteMaxAbs} on O*NET. On DBpedia, whose bank of \res{bankDbpedia} articles makes retrieval alone strong, the
gain is at most \res{dDbpediartdVvoteMaxAbs} points.

Table~\ref{tab:allarms} gives the coverage of every setting, the share of queries whose gold label is among the
candidates, and its accuracy. The adaptive set covers \res{covTauBanking}\%, \res{covTauClinc}\% and
\res{covTauDbpedia}\% of queries on the three smaller label sets and \res{covTauOnet}\% on O*NET, with
\res{ktauBanking}, \res{ktauClinc}, \res{ktauDbpedia} and \res{ktauOnet} candidates on average: it keeps the menu
short when retrieval scores are concentrated on a few labels and longer when they are more dispersed. With $K{=}5$ only
\res{cov5Onet}\% of O*NET queries keep their gold label, and Jev-1.14 drops from \res{accOnetJevknntau}\% to
\res{accOnetJevknnFive}\%.

With a hash-selected tenth of each bank, AdaMenu outperforms the hierarchical baseline on all
\res{allHashVtreeN} comparable pairs. Relative to the full bank, the largest accuracy loss on the three smaller
label sets is \res{withinHashVfullMinAbs} points; O*NET shows larger losses (Appendix~\ref{app:bank}).

\section{Conclusion}
\label{sec:conclusion}

Large label spaces pose challenges to decision accuracy as well as scalability: expanding a menu can impair
classification and increase input cost even before reaching the model's option limit. AdaMenu addresses these
challenges by constructing compact, query-adaptive candidate menus and making a single decision-model request.
It improves accuracy over TypeSafe's beam-search baseline on all comparable model and label-set pairs.
On the three smaller label sets, it achieves competitive accuracy with fewer input tokens than the full menu.
It also enables classification when the full menu exceeds the option limit, including settings where
the hierarchical baseline is unavailable.

Accuracy remains limited by candidate coverage and the model's ability to distinguish the retained labels.
Improving candidate construction and developing mechanisms for detecting and recovering from candidate
omissions are directions for future work.

\section*{Limitations}

Jev results correspond to the API snapshot in Appendix~\ref{app:models}; open-model results come from our
serving-code re-implementations. Hosted API latency also depends on provider load.

BANKING77 is evaluated on a held-out part of its training split, not its official test split. For O*NET,
the bank and queries come from the same occupational database and may contain close paraphrases. Bootstrap
intervals are computed by resampling queries with models held fixed, without adjustment for multiple comparisons.

Candidate coverage bounds accuracy, and the method has no explicit rejection or recovery mechanism.
On O*NET, a hash-selected tenth of the bank lowers AdaMenu accuracy by
\res{dOnetHashVFullMaxAbs}--\res{dOnetHashVFullMinAbs} percentage points relative to the full bank
(Appendix~\ref{app:bank}). This subsampling changes both label coverage and the examples available within
retained labels.

We evaluate English data with one retriever (BM25), label names as option descriptions and the supplied label
trees. AdaMenu uses a labeled example bank that the full-menu classifier and hierarchical baseline do not
access. We study inference with existing decision models and do not compare with classifiers trained on
the bank, which can be strong when labeled training examples are plentiful \citep{rafe2026benchmarking}.

\bibliography{refs}

\appendix
\renewcommand{\dbltopfraction}{0.9}
\renewcommand{\dblfloatpagefraction}{0.9}
\renewcommand{\textfraction}{0.08}
\raggedbottom
\begin{table*}[t]
\centering\small\setlength{\tabcolsep}{4.2pt}

\begin{tabular}{@{}l rrrr rrrr rrrr@{}}
\toprule
 & \multicolumn{4}{c}{CLINC150} & \multicolumn{4}{c}{DBpedia} & \multicolumn{4}{c}{O*NET} \\
\cmidrule(lr){2-5}\cmidrule(lr){6-9}\cmidrule(l){10-13}
Model & Full & Tree & \textit{leaf$|$top} & AdaMenu & Full & Tree & \textit{leaf$|$top} & AdaMenu & Full & Tree & \textit{leaf$|$top} & AdaMenu \\
\midrule
Jev-1.14 & 97.4 & 70.1 & 95.8 & 97.9 & 98.6 & 90.2 & 82.1 & 98.7 & -- & 69.9 & 62.7 & 84.0 \\
Jeff-1.1-Qwen-4B & 96.9 & 65.2 & 95.2 & 97.7 & 98.2 & 85.5 & 82.3 & 98.6 & -- & 64.1 & 57.6 & 81.6 \\
StartLux-4B & 99.0 & 76.4 & 98.7 & 98.8 & 98.4 & 88.5 & 80.8 & 98.7 & 84.5 & 64.8 & 55.7 & 82.6 \\
Eikos-4B & 92.6 & 59.4 & 90.9 & 96.5 & 95.5 & 71.8 & 78.4 & 97.9 & 78.4 & 58.4 & 50.1 & 79.6 \\
Eikos-27B & 98.2 & 64.5 & 94.3 & 97.8 & 97.8 & 86.9 & 82.7 & 98.6 & 84.5 & 68.8 & 58.1 & 82.8 \\
Jeff-1.0-Large & -- & 69.0 & 97.0 & 97.4 & -- & -- & -- & 98.7 & -- & -- & -- & 82.7 \\
\bottomrule
\end{tabular}

\caption{First-step routing per model (\%). Full (the full menu, read in one pass or split) / AdaMenu: share of the
method's predictions under the gold label's top-level node. Tree: share of queries whose greedy first step picks that node. \textit{leaf$|$top}: accuracy of
greedy descent among the queries it routed correctly. --: menu not accepted.}
\label{tab:routing}
\end{table*}

\begin{table*}[t]
\centering\small

\begin{tabular}{@{}ll rrrrrrrr@{}}
\toprule
Data & Model & Full & Split & Greedy & Beam-3 & $K{=}5$ & $K{=}10$ & $K{=}20$ & $\tau$ \\
\midrule
BANKING77 & \textit{Coverage} & & & & & 96.0 & 98.1 & 99.3 & 99.0 \\
 & Jev-1.14 & 79.3 & -- & -- & -- & 80.9 & 80.3 & 80.4 & 80.1 \\
 & Jeff-1.1-Qwen-4B & -- & 78.3 & -- & -- & 80.4 & 80.5 & 78.2 & 80.8 \\
 & StartLux-4B & -- & 89.2 & -- & -- & 87.8 & 88.9 & 89.7 & 89.6 \\
 & Eikos-4B & 70.8 & -- & -- & -- & 76.6 & 76.1 & 73.1 & 76.4 \\
 & Eikos-27B & 77.1 & -- & -- & -- & 78.7 & 79.0 & 76.6 & 79.6 \\
 & Jeff-1.0-Large & -- & -- & -- & -- & 81.5 & 80.4 & -- & -- \\
\midrule
CLINC150 & \textit{Coverage} & & & & & 95.4 & 97.3 & 98.5 & 98.2 \\
 & Jev-1.14 & 92.3 & -- & 67.2 & 69.9 & 91.2 & 92.1 & 92.2 & 93.0 \\
 & Jeff-1.1-Qwen-4B & -- & 90.8 & 62.1 & 68.8 & 90.7 & 91.1 & 90.8 & 92.2 \\
 & StartLux-4B & -- & 96.3 & 75.4 & 79.2 & 93.4 & 94.7 & 95.7 & 95.8 \\
 & Eikos-4B & -- & 81.7 & 54.0 & 60.0 & 87.6 & 87.2 & 86.9 & 88.6 \\
 & Eikos-27B & 92.7 & -- & 60.9 & 65.8 & 90.2 & 90.9 & 91.4 & 92.2 \\
 & Jeff-1.0-Large & -- & -- & 66.9 & 72.0 & 91.7 & 92.8 & -- & -- \\
\midrule
DBpedia & \textit{Coverage} & & & & & 98.8 & 99.4 & 99.7 & 99.2 \\
 & Jev-1.14 & 93.0 & -- & 74.0 & 75.1 & 93.0 & 92.7 & 92.7 & 93.3 \\
 & Jeff-1.1-Qwen-4B & -- & 91.8 & 70.3 & 72.0 & 92.7 & 92.2 & 92.3 & 93.1 \\
 & StartLux-4B & -- & 91.5 & 71.5 & 73.1 & 92.1 & 92.4 & 91.8 & 92.8 \\
 & Eikos-4B & -- & 83.9 & 56.3 & 56.8 & 89.9 & 89.0 & 88.3 & 91.0 \\
 & Eikos-27B & -- & 91.5 & 71.8 & 71.2 & 92.6 & 92.1 & 92.1 & 93.2 \\
 & Jeff-1.0-Large & -- & -- & -- & -- & 92.5 & 92.5 & -- & 89.6 \\
\midrule
O*NET & \textit{Coverage} & & & & & 73.7 & 81.0 & 85.9 & 86.2 \\
 & Jev-1.14 & -- & -- & 43.7 & 44.4 & 64.1 & 67.9 & 70.2 & 70.1 \\
 & Jeff-1.1-Qwen-4B & -- & -- & 36.9 & 37.9 & 62.7 & 65.4 & 66.3 & 66.2 \\
 & StartLux-4B & -- & 69.5 & 36.1 & 38.2 & 63.5 & 66.6 & 67.4 & 67.4 \\
 & Eikos-4B & -- & 57.2 & 29.2 & 29.7 & 59.4 & 61.6 & 62.7 & 62.0 \\
 & Eikos-27B & -- & 68.0 & 40.0 & 40.1 & 62.7 & 65.9 & 67.3 & 66.9 \\
 & Jeff-1.0-Large & -- & -- & -- & -- & 64.0 & 68.1 & -- & -- \\
\bottomrule
\end{tabular}

\caption{Accuracy (\%) of every method and model. Full, Split: all labels in one request, read in one pass (Full) or
split by the model's serving code into groups read in several passes (Split). Greedy, Beam-3: variants of the
hierarchical baseline. $K$, $\tau$: AdaMenu with $K \in \{5, 10, 20\}$ candidates or adaptive. \textit{Coverage}: share of queries
whose gold label is among the candidates of each AdaMenu setting. --: not available (no supplied label tree, or more options than the
model accepts).}
\label{tab:allarms}
\end{table*}

\begin{table*}[t]
\centering\footnotesize\setlength{\tabcolsep}{4pt}

\begin{tabular}{@{}ll rrr rrrrrr@{}}
\toprule
 & & & & & \multicolumn{6}{c}{AdaMenu accuracy} \\
\cmidrule(l){6-11}
Data & Bank & No ex. & Cov. & Top-1 & Jev & Jeff-1.1 & StartLux & Eikos-4B & Eikos-27B & Jeff-1.0-L \\
\midrule
BANKING77 & full & 0 & 99.0 & 78.7 & 80.1 & 80.8 & 89.6 & 76.4 & 79.6 & 80.4 \\
 & 10\% & 0 & 97.2 & 55.2 & 78.6 & 77.8 & 88.2 & 73.0 & 77.2 & 76.0 \\
 & 10\%, strat. & 0 & 97.3 & 61.0 & 79.3 & 78.2 & 87.7 & 71.3 & 77.1 & 76.3 \\
\midrule
CLINC150 & full & 0 & 98.2 & 80.6 & 93.0 & 92.2 & 95.8 & 88.6 & 92.2 & 92.8 \\
 & 10\% & 0 & 94.2 & 60.3 & 88.9 & 87.8 & 92.1 & 83.9 & 87.7 & 87.2 \\
 & 10\%, strat. & 0 & 94.8 & 67.0 & 89.9 & 88.9 & 92.8 & 84.5 & 88.9 & 87.5 \\
\midrule
DBpedia & full & 0 & 99.2 & 91.1 & 93.3 & 93.1 & 92.8 & 91.0 & 93.2 & 92.5 \\
 & 10\% & 0 & 99.2 & 85.5 & 93.1 & 92.6 & 92.2 & 90.2 & 92.8 & 92.6 \\
 & 10\%, strat. & 0 & 99.2 & 85.5 & 92.8 & 92.6 & 92.2 & 90.0 & 92.9 & 92.5 \\
\midrule
O*NET & full & 2 & 86.2 & 42.6 & 70.1 & 66.2 & 67.4 & 62.0 & 66.9 & 68.1 \\
 & 10\% & 114 & 54.2 & 17.7 & 46.7 & 44.4 & 44.6 & 42.3 & 44.8 & 41.0 \\
 & 10\%, strat. & 2 & 59.4 & 19.6 & 50.8 & 48.4 & 49.0 & 45.8 & 48.5 & 44.4 \\
\bottomrule
\end{tabular}

\caption{AdaMenu with the full bank, a hash-selected tenth and a stratified tenth (every label with an example in the full bank keeps at least one).
No ex.: labels without any example. Cov.: coverage of the adaptive candidates (\%). Top-1: accuracy of the top-1 kNN
candidate (\%). AdaMenu accuracy (\%): adaptive menu sizes, except $K{=}10$ for Jeff-1.0-Large.}
\label{tab:bank}
\end{table*}

\section{Routing, candidate sets and every method}
\label{app:routing}

Table~\ref{tab:routing} gives the first-step routing per model (\S\ref{sec:why}), and
Table~\ref{tab:allarms} the accuracy of every method, with the coverage of every AdaMenu setting
(\S\ref{sec:parts}).

\section{Models and requests}
\label{app:models}

For Jev-1.14, all requests went through OpenRouter's decision endpoint in November 2026, with the query as
the request state and one \Choice{} question per request, except that beam search sends the questions of all beam
nodes of a level in one request. The option keys are the label names, and each option's description is its key.
Responses report the served snapshot \texttt{typesafe/jev-1.14-20261014}.

Jeff-1.1-Qwen-4B, Jeff-1.0-Large, StartLux-Decision-v1.1-4B-Instant, Eikos-4B and Eikos-27B
run on our GPUs. For each we re-implemented the request handling of the vendor's public serving code: the prompt
built from a TypeSafe-style request, the readout (a decision head or the language-model logits of the option labels),
and the handling of menus longer than one read (Table~\ref{tab:models}): Jeff-1.1 reads groups of 16 options and the
group winners in a final read, StartLux-4B runs a knock-out over groups of 25, the Eikos models run a tournament over
reads of 100 (4B) or 160 (27B) options, and Jeff-1.0-Large rejects more than 16 options. We checked the
re-implementations on OpenDecisions and compared their scores with published reference scores where available.

Jeff-1.0-Large uses the fixed-size $K{=}10$ variant because it accepts at most 16 options.
The other models use adaptive menus in the main comparison. The adaptive upper bound of $K_{\max}=26$
exceeds the single-pass capacities of Jeff-1.1-Qwen-4B and StartLux-4B; Jeff-1.1-Qwen-4B averages
\res{readsOnetJeffknntau} reads per query on O*NET.

On 2026-11-09 we sent each of the \res{probeModels} hosted decision models then served
on OpenRouter one request whose single \Choice{} question lists \res{probeOptions} options (Figure~\ref{fig:teaser}c).
It was rejected by \res{probeRejectedOf} with a client error, and the messages of \res{probeCapN} state a limit of
\res{probeCap} options per question.

\section{Candidate menu construction}
\label{app:candidates}

BM25 uses lowercased tokens matching \texttt{[a-z0-9]+} and adjacent word pairs, hashed into $2^{22}$ features;
its parameters are given in \S\ref{sec:setup}. Bank examples are ordered by descending retrieval score,
with ties broken by their index in the bank. Label scores $r_x(y)$ sum the positive scores of the first $M$ examples.
Labels with positive scores are ordered by decreasing $r_x(y)$; ties follow the labels' first occurrence
in the ranked examples.

The ordered list $\pi_x$ initially contains all labels with positive scores. If it has fewer than
$K_{\max}$ labels, we scan examples after the first $M$ in retrieval order and append each previously
unlisted label until the list reaches $K_{\max}$ or the examples are exhausted. If it is still shorter,
we append unlisted labels in the dataset's fixed full-label order, stopping at $K_{\max}$ or when no labels
remain. These appended labels contribute zero to the score mass, since $r_x(y)$ is determined only by the
first $M$ examples. The resulting list has length $m_x$ before the final menu is truncated; it can exceed
$K_{\max}$ if more than $K_{\max}$ labels have positive scores.

For positive total score $R_x$, the adaptive rule first finds the shortest prefix containing a fraction
$\tau$ of that score, then applies the bounds $K_{\min}$ and $K_{\max}$ and the number of
available labels, as in \S\ref{sec:method}. The upper bound can leave the final menu with less than the target
fraction of score mass. If $R_x{=}0$, we use $K(x)=\min(K_{\max},m_x)$, so the menu contains $K_{\max}$ labels
when that many are available, and all available labels otherwise. Fixed-size variants take the first $K$
labels of the same ordered list, or all available labels if fewer than $K$ exist.

\section{Sensitivity to the size of the bank}
\label{app:bank}

The coverage of the candidate set depends on the bank. We rerun AdaMenu (BM25, same settings) with a tenth of each bank,
selected in two ways: by a hash of the row, so that a label can lose all its examples, as in a small real bank; and
stratified, so that every label with an example in the full bank keeps at least one (Table~\ref{tab:bank}). The queries, the full menu and the
hierarchical baseline are unchanged.

With the hash-selected tenth, AdaMenu outperforms the hierarchical baseline on all
\res{allHashVtreeN} model and label-set pairs, by \res{allHashVtreeMinAbs} to \res{allHashVtreeMaxAbs} points (all
significant); with the stratified tenth, by \res{allStratVtreeMinAbs} to \res{allStratVtreeMaxAbs}.

On the three smaller label sets, every label retains examples in both reduced banks. With the hash-selected
tenth, accuracy losses relative to the full bank are
\res{dBankingHashVFullMaxAbs} to \res{dBankingHashVFullMinAbs} points on BANKING77, \res{dClincHashVFullMaxAbs} to
\res{dClincHashVFullMinAbs} on CLINC150, and at most \res{dDbpediaHashVFullMinAbs} on DBpedia, whose tenth still
holds \res{bankRowsDbpediaHash} articles; with the stratified tenth, \res{dBankingStratVFullMaxAbs} to
\res{dBankingStratVFullMinAbs}, \res{dClincStratVFullMaxAbs} to \res{dClincStratVFullMinAbs} and at most
\res{dDbpediaStratVFullMinAbs}. Against the full menu, AdaMenu with the hash-selected tenth is not significantly worse on
\res{withinHashVflatNotNeg} of \res{withinHashVflatN} pairs while reading a fraction of its tokens.

O*NET has \res{labelsOnet} occupations, but the
hash-selected tenth of its bank contains only \res{bankRowsOnetHash} titles: \res{lostLabelsOnetHash} occupations keep no example at all, which affects
\res{lostQueriesOnetHash} queries, and coverage falls from \res{covTauOnetFull}\% to \res{covTauOnetHash}\%. AdaMenu then
loses \res{dOnetHashVFullMaxAbs} to \res{dOnetHashVFullMinAbs} points. The stratified tenth, which retains at least
one example for every label represented in the full bank, recovers \res{dOnetStratVHashMinAbs} to
\res{dOnetStratVHashMaxAbs} points, almost all of it on
those queries (\res{dOnetStratVHashLostMin} to \res{dOnetStratVHashLostMax} points there, \res{dOnetStratVHashKeptMin}
to \res{dOnetStratVHashKeptMax} on the rest). Its accuracy remains below that obtained with the full bank for
every model.

\section{Reproducibility}
\label{app:repro}

All numbers, tables and figures are computed from stored per-query records: the prediction, token counts, price,
latency and served snapshot of every request to Jev-1.14, and the predictions, prompt tokens and reads per query of
the open models.

\end{document}